\documentclass{article} 
\usepackage{iclr2024_conference,times}

\usepackage{amsmath,amsfonts,bm}

\def\eqref#1{equation~\ref{#1}}

\def\1{\bm{1}}

\DeclareMathAlphabet{\mathsfit}{\encodingdefault}{\sfdefault}{m}{sl}
\SetMathAlphabet{\mathsfit}{bold}{\encodingdefault}{\sfdefault}{bx}{n}

\usepackage{amsmath}
\usepackage{amssymb}
\usepackage{mathtools}
\usepackage{amsthm}
\usepackage{subcaption}
\usepackage{wrapfig}
\usepackage{tikz}

\usepackage{hyperref}
\usepackage{url}
\newtheorem{theorem}{Theorem}

\newtheorem{lemma}[theorem]{Lemma}

\newtheorem{manuallemmainner}{Lemma}
\newenvironment{manuallemma}[1]{%
  \renewcommand\themanuallemmainner{#1}%
  \manuallemmainner
}{\endmanuallemmainner}

\title{Direct Alignment of Draft Model for Speculative Decoding with Chat-Fine-tuned LLMs}

\author{Raghavv Goel\thanks{Correspondence to: \texttt{$\{$raghgoel, mingul, clott$\}$@qualcomm.qti.com} \\ Qualcomm AI Research is an initiative of Qualcomm Technologies, Inc.}, Mukul Gagrani, Wonseok Jeon, Junyoung Park, Mingu Lee, Christopher Lott 
\\
Qualcomm AI Research
\\
}

\iclrfinalcopy 
\begin{document}

\maketitle

\begin{abstract}

Text generation with Large Language Models (LLMs) is known to be memory bound due to the combination of their auto-regressive nature, huge parameter counts, and limited memory bandwidths, often resulting in low token rates. Speculative decoding has been proposed as a solution for LLM inference acceleration. 
However, since draft models are often unavailable in the modern open-source LLM families, e.g., for Llama 2 7B, training a high-quality draft model is required to enable inference acceleration via speculative decoding.
In this paper, we propose a simple draft model training framework for direct alignment to chat-capable target models. With the proposed framework, we train Llama 2 Chat Drafter 115M, a draft model for Llama 2 Chat 7B or larger, with only 1.64\% of the original size. 
Our training framework only consists of pretraining, distillation dataset generation, and finetuning with knowledge distillation, with no additional alignment procedure. For the finetuning step, we use instruction-response pairs generated by target model for distillation in plausible data distribution, and propose a new Total Variation Distance++ (TVD++) loss that incorporates variance reduction techniques inspired from the policy gradient method in reinforcement learning. Our empirical results show that Llama 2 Chat Drafter 115M with speculative decoding achieves up to 2.3 block efficiency and 2.4$\times$ speed-up relative to autoregressive decoding on various tasks with no further task-specific fine-tuning.
\end{abstract}
\vspace{-0.15in}
\section{Introduction}
\vspace{-0.1in}




Large language models (LLMs) have become universal and versatile tools in our daily life thanks to their impressive results and superior performance in a wide range of domains \citep{achiam2023gpt, anil2023palm, roziere2023code}. There has been increasing demands for running edge LLMs directly on user devices for numerous reasons such as privacy, security, cost reduction, reliability, and personalization. However, LLMs naturally generates tokens in an auto-regressive manner which is an inherently slow process due to memory bandwidth bottleneck.
Speculative decoding (SD) has been proposed to mitigate the inference speed bottleneck and accelerate LLM inference by introducing a smaller draft model to predict the output of the large target model\citep{leviathan2023fast, chen2023accelerating}. In each iteration of SD, the draft model generates a sequence of tokens and the target model accepts a sub-sequence of the draft tokens using a modified rejection sampling criteria. It has been shown that SD can provide up to 2-3$\times$ speedup in LLM inference without any loss in text generation quality.

Despite the promising performance of SD, high-quality small language models that can serve as draft models are often unavailable, since the smallest members of modern open-source LLM families, such as Llama 2 \citep{touvron2023llama}, Falcon \citep{refinedweb}, and Baichuan \citep{Baichuan} are at the scale of several billions of parameters. Models with this scale are either considered as target models on edge devices or too large as draft models to get meaningful inference speed improvement with SD. Enabling SD with these large models on small devices require a much smaller draft model which is well aligned with the target model. When it comes to training a draft model for a given target model, a practical challenge is that the original dataset used for training the target model may not be provided that prohibits following the same training procedure of the target model as a natural way of mimicking the target model behavior.
Recent works for draft model training has taken inspiration from knowledge distillation \citep{hinton2015distilling} where student models were trained using token-level distillation \citep{kim2016sequence} which was extended to distribution-level distillation \citep{zhou2023distillspec, agarwal2023gkd, wen2023f, lin2020autoregressive} when the entire teacher model distributions are accessible. These methods use either KL-divergence (forward, backward or Jenson-Shannon) or total variation distance for training the student model, focusing only on improving student model generation instead of improving SD performance \citep{agarwal2023gkd, wen2023f}. 

In this work, we propose a framework for draft model training for speculative decoding. Our training pipeline consists of three phases 
: 1) pre-training; training a randomly initialized draft model on a large open source text corpus to gain the general language modeling capabilities. 2) generation of a distillation dataset by having the target model generate responses to various instructions. Since the goal of the draft model is to mimic the target model we first generate the plausible data distribution using open source instruction fine-tuning datasets. 3) Fine-tuning on the distillation dataset for aligning draft and target model behaviors with target model in the loop. In addition, we propose a novel distillation loss, Total Variation Distance++ (TVD++) loss, by making connections between the acceptance of drafts in SD and policy gradient method in reinforcement learning.

With this framework, we train a draft model for Llama 2-Chat-7B model \citep{touvron2023llama} of size 115M which is only 1.64\% of the size of the target model, Llama 2-Chat-Drafter-115M. We evaluate our draft model on various tasks like open-ended generation on Databricks-dolly-15k dataset \citep{DatabricksBlog2023DollyV2} and text summarization on XSum \citep{narayan2018don}, CNN/DailyMail datasets \citep{nallapati2016abstractive}. We show that our trained model can obtain up to 2.4$\times$ speed-up with speculative decoding over auto-regressive inference demonstrating the effectiveness of our training pipeline. Additionally, our proposed loss outperforms commonly used distillation losses: KLD and TVD. 

\vspace{-0.1in}
\section{Method}
\vspace{-0.1in}
Our draft model training pipeline involves three steps 
with the goal of aligning the output of draft model with that of the target language model in the context of speculative decoding process, as opposed to having the draft model as a standalone language model. 
\vspace{-0.1in}
\subsection{Draft model pretraining}
\vspace{-0.1in}
    A draft model is required to have the same tokenizer as the corresponding target model with the same or a subset of vocabulary.
    Some language model families sharing the same tokenizer have wide ranges of model sizes where we can pick draft models at a desired sizes, e.g., Pythia~\citep{biderman2023pythia}. However, this is not the case more often, e.g., Llama 2~\citep{touvron2023llama}, where the smallest is still too large. In such cases, we train a draft language model from scratch with the regular next token prediction loss in order to have a good language model to be aligned with the target model more effectively. It is possible to incorporate initialization techniques such as Weight Subcloning~\citep{samragh2023weight} to expedite this step.
    While it may be trivial, in our empirical observations, pretrained draft models show significantly better alignment to target model, compared to those without pretraining. 

\vspace{-0.075in}
\subsection{Distillation dataset generation}
\vspace{-0.075in}
\label{subsec:generating_distillation_data}
In language modeling, there are virtually infinite number of possible input data since the same input token can produce different outputs according to the conditions made by the specific contexts, i.e., different prompts and past generations, in the causal attention mechanism. To mimic the target behavior in realistic contexts, it is crucial to have data samples that are plausible in the target model generation. While one convenient way is follow the training procedure of the target model including the same instruction and alignment datasets, availability of such datasets and reward models are not always guaranteed.
To mitigate such challenges and simplify the training procedure, we generate a {\it distillation dataset} by using seed instructions from publicly available datasets and letting the target model generate a diverse set of responses in various configuration, e.g., temperature, top-p, and system prompts. Note that unlike \cite{zhou2023distillspec, agarwal2023gkd}, we only use target model for generating responses as opposed to using the pre-trained draft model. This step can be seen as a data-level distillation.


\vspace{-0.05in}
\subsection{Draft model finetuning via knowledge distillation}
\vspace{-0.05in}
We then finetune the pretrained draft model using dataset generated as in \ref{subsec:generating_distillation_data}. Knowledge distillation comes in two different flavors (a) black-box distillation \citep{kim2016sequence} where the target model is not accessible and only the token generated by target model can be used for distillation, and (b) white-box distillation, where we have access to the entire target output distribution. White-box optimization provides much stronger learning signals given the loss is optimized over distribution as opposed to a single token label, which corresponds to our case.

The white-box distillation can be thought as a distribution matching problem, where the target model ($\mathcal{M}_{q}$) output distribution ($q(y|x)$) for a given input text ($x$) is the desired distribution for the draft model ($\mathcal{M}_{p}(\theta)$) output distribution ($p_{\theta}(y|x)$). The training objective can be defined using a distance metric between distributions such as Kullback–Leibler Divergence (KLD) and its variants: backward KLD, Jenson-Shannon Divergence \citep{agarwal2023gkd}, or Total Variation Distance (TVD). Theoretical analysis based on SD (\textit{Corollary 3.6} in \citep{leviathan2023fast}) has shown that minimizing TVD is equivalent to maximizing acceptance-rate, the true objective of improving SD performance.  In this paper, we further 
build on TVD using techniques from reinforcement learning. We first show that the gradient step taken when optimizing TVD loss is equivalent to the policy-gradient step for reward maximization in reinforcement learning. Based on this, we can bring tools from reinforcement learning such as variance reduction techniques for improving knowledge distillation loss in LLMs \citep{korbak2022reinforcement}.

\begin{lemma}
    The gradient of total variation distance $\mathrm{TVD}(p_\theta, q)$ between the draft model $p_\theta$ and the target model $q$ w.r.t. the draft model parameter $\theta$ is equal to 
    $
        \nabla_\theta 
        \mathrm{TVD}(p_\theta, q)
        =
        \mathbb{E}_{X\sim p_\theta}
        \left[
            \nabla_\theta 
            \log p_\theta(X)(-r(X))
        \right]
    $
    for $r(x):=\mathbb{I}\{q(x) > p_{\theta}(x)\}$, where $\mathbb{I}\{q(x) > p_{\theta}(x)\}$ is an indicator function, which is equal to $1$ if $q(x)>p_{\theta}(x)$, or $0$, otherwise.
    \label{lemma:TVD_RL}
\end{lemma}
\vspace{-0.15in}
\begin{proof}
See the proof in Appendix~\ref{subec:lemma_proof}.    
\end{proof}
\vspace{-0.1in}

Using Lemma \ref{lemma:TVD_RL}, we can use variance reduction techniques from reinforcement learning \citep{schulman2015high} to formulate a new loss. Specifically, for our purpose, we use the advantage normalization which normalizes the reward and propose {\it Total Variation Distance++ (TVD++)}. The normalization leads to negative rewards as opposed to $0$ reward which we believe could lead to better learning signals for draft model fine-tuning. The stochastic gradient of TVD++ for a sample of $n$ tokens is given as:
\begin{equation}
    \nabla_{\theta}\mathrm{TVD}^{++}(p_{\theta}, q)=\frac{1}{n}\sum_{i=1}^{n}\nabla_{\theta}\log p_{\theta}(x_{i})\left(\frac{r(x_{i})-\mu}{\sigma}\right),
\end{equation}
where $\mu=\frac{1}{n}\sum_{i=1}^{n}r(x_{i})$ and $\sigma^2=\frac{1}{n}\sum_{i=1}^{n}(r(x_{i})-\mu)^{2}$ are the mean and variance for the $n$ samples (tokens from vocabulary) for $r$ in \textbf{Lemma~\ref{lemma:TVD_RL}}. Note that we use the entire distribution of target, and the mean, variance are computed over the input sequences and the entire vocabulary. 

\section{Experiments}
\textbf{Draft Model Configuration.}
We design a draft model by using smaller number of layers and hidden intermediate dimensions than those of Llama 2 7B Chat, our target model, based on the same network architecture (See Appendix \ref{subsec:model_config} for detailed configurations.).
The size of our draft model  is \textbf{115M}, which is only \textbf{1.64\%} than the size of the target model \textbf{7B} to achieve negligible inference overheads even on edge device, e.g., mobile phones or laptops.

\textbf{Training Datasets.}
Since Llama 2 pretraining dataset is not available, our draft model is pretrained on a 600B-token English language dataset that is curated from publicly available datasets. using the next token prediction objective. 

For generating the {\it distillation dataset}, we take seed instructions from \textbf{OIG-small-chip2}\citep{OIG} and \textbf{OpenAssistant} \citep{kopf2023openassistant} to collect the target model responses sampled with various temperatures $\{0, 0.3, 0.7, 1.0\}$ and top-${p}=0.95$ for sample diversity (temperature=$0$ corresponds to greedy decoding).

\textbf{Draft fine-tuning.}
To compare the effectiveness of our {\it TVD++}, we use different loss functions to fine-tune the draft output distribution to align with target output distribution: (i) \textbf{KLD} (ii) \textbf{TVD} and (iii) \textbf{TVD++}. In this step, we mix the distillation dataset with the pretraining dataset at 9:1 ratio in each training batch for regularization effect.

\textbf{Evaluation metrics and tasks.}
We evaluate our draft model by measuring the (1) \textbf{block efficiency} ($\tau$): average number of tokens generated per block (or target model run), for a block of size $\gamma$ and input $x$, the maximum value of $\tau(x)$ can be $\gamma + 1$, (2) memory-bound speed-up (\textbf{MBSU}): hypothetical speed-up achieved by SD for a given block efficiency $\tau(x)$ and a relative latency $c$ defined as ratio between number of parameters of draft to target model, i.e., $\mathrm{MBSU}(x):=\frac{c\tau(x)}{c\gamma + 1}$, and (3)\textbf{Token Rate}: number of tokens generated per second, in our case we measure the ratio of the SPD token-rate to auto-regressive token-rate, a ratio $>1$ implies higher number of tokens were generated per second for SPD. We measure these metrics on a variety of tasks using different block lengths $\gamma$ in $\{3, 5\}$. The evaluation tasks include: (a) open-ended text generation \citep[databricks-\textbf{Dolly}-15k]{DatabricksBlog2023DollyV2}, (b) extreme summarization \citep[\textbf{XSum}]{narayan2018don}, 
and (c) news summarization \citep[\textbf{CNN-dailymail}]{nallapati2016abstractive}. For open-ended text generation we random-sample with \textbf{temperature}$=0.6$ and \textbf{top-$p$}$=0.9$ for both draft model and target model, while we greedy-sample for the rest.

\begin{figure}
\vspace{-0.45in}
\centering
\includegraphics[width=\linewidth]{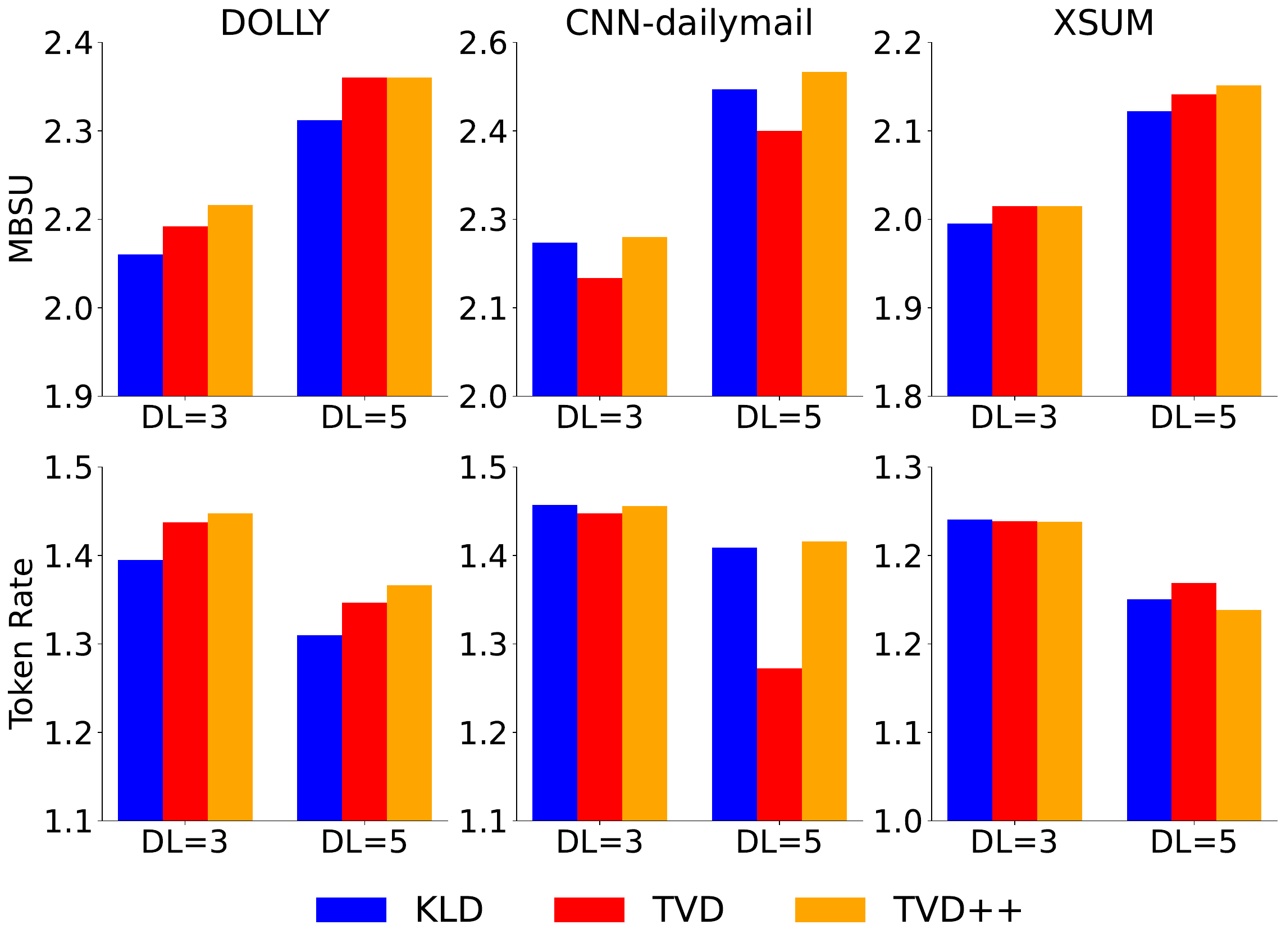}
\caption{
Draft models evaluated on Memory-Bound Speed-Up (MBSU) and token-rate (realtive to auto-regressive generation) for multiple tasks (Dolly, CNN-dailymail, XSum) with draft lengths (3, 5) and training losses (KLD, TVD, TVD++);
}
\label{fig:mbsu_all}
\end{figure}

\begin{figure}
\vspace{-0.2in}
\centering
\includegraphics[width=\linewidth]{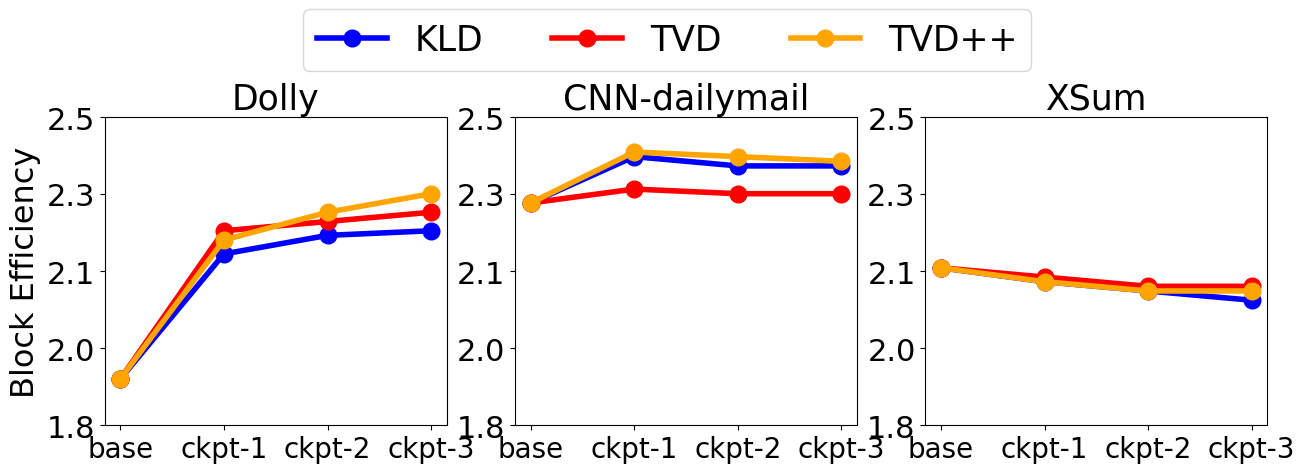}
\caption{Draft models evaluated on block efficiency ($\gamma=3$) over different checkpoints (ckpt) across fine-tuning stage, showing improvement over the base draft model, for multiple tasks (Dolly, CNN-DM, XSum) and training losses (KLD, TVD, TVD++)}
\label{fig:token_trend_DL=3}
\vspace{-0.1in}
\end{figure}

\textbf{Results.}
Figure \ref{fig:mbsu_all} shows MBSU for draft models fine-tuned using different losses, with $\gamma= 3$ or $5$. Our proposed TVD++ loss either outperforms the other two losses or performs on par with the best on all the tasks, showing its effectiveness for improving draft model fine-tuning.
Furthermore, Figure \ref{fig:token_trend_DL=3} shows the improvement in block-efficiency with $\gamma=3$ across different training checkpoints during fine-tuning stage. For open-ended text generation task (Dolly), we observe block efficiency increases by around $\sim 21\%$ with more fine-tuning over the base draft model, while block efficiency improvement is also observed for news summarization task (CNN-dailymail) from $2.29$ for base draft to $2.4$ for fine-tuned draft showing the efficacy of fine-tuning.  

\section{Conclusion}
In this work, we proposed a framework to train a draft model with direct alignment to a finetuned target LLM for speculative decoding. Our training pipeline consists of pre-training, generating a distillation dataset from the target LLM and fine-tuning the draft model on the distillation dataset with target LLM in the training loop. In addition, we further proposed a new loss building on TVD inspired from policy gradient in RL which outperforms both KLD and TVD in fine-tuning.

\bibliography{iclr2024_conference}
\bibliographystyle{iclr2024_conference}

\clearpage
\appendix
\section{Appendix}

\subsection{Proof of Lemma \ref{lemma:TVD_RL}}
\label{subec:lemma_proof}

\begin{manuallemma}{\ref{lemma:TVD_RL}}
    The gradient of total variation distance $\mathrm{TVD}(p_\theta, q)$ between the draft model $p_\theta$ and the target model $q$ w.r.t. the draft model parameter $\theta$ is equal to 
    \begin{align*}
        \nabla_\theta 
        \mathrm{TVD}(p_\theta, q)
        =
        \mathbb{E}_{X\sim p_\theta}
        \left[
            \nabla_\theta 
            \log p_\theta(X)(-r(X))
        \right]
    \end{align*}
    for $r(x):=\mathbb{I}\{q(x) > p_{\theta}(x)\}$, where $\mathbb{I}\{q(x) > p_{\theta}(x)\}$ is an indicator function, which is equal to $1$ if $q(x)>p_{\theta}(x)$, or $0$, otherwise.
    
\end{manuallemma}
\begin{proof}
    Let the draft and target model output distributions be defined as $p_{\theta}(x)$ and $q(x)$, respectively, where $\theta$ is the trainable draft model parameter and $x \in \{1, \dots,|\mathcal{V}|\}$, where $\mathcal{V}$ is a set of tokens, i.e., vocabulary. We ignore the conditioning on input prompt here for ease of understanding.
    From \cite{leviathan2023fast}, TVD can be represented as
    \begin{align}
        \mathrm{TVD}(p_{\theta}, q) =& \frac{1}{2}\sum_{x}|q(x)-p_{\theta}(x)|= 1 - \sum_{x} \min (p_{\theta}(x), q(x)).
        \label{eq:A:1}
    \end{align}
    Then, one can derive the gradient of RHS in \eqref{eq:A:1} w.r.t. $\theta$ as
    \begin{align}
        \nabla_{\theta} \mathrm{TVD}(p_{\theta}, q) &= \nabla_{\theta}\big(1 - \sum_{x} \min (p_{\theta}(x), q(x)) \big)\nonumber \\
        &= -\sum_{x}\mathbb{I}\{q(x) > p_{\theta}(x)\}\nabla_{\theta}p_{\theta}(x) \nonumber\\
        &= \sum_{x}p_{\theta}(x)(-\mathbb{I}\{q(x) > p_{\theta}(x)\})\nabla_{\theta}\log p_{\theta}(x) \label{eq:A:2}\\
        &=\mathbb{E}_{x\sim p_{\theta}(x)}\left[
        \nabla_{\theta}\log p_{\theta}(x)
        (-\mathbb{I}\{q(x) > p_{\theta}(x)\})
        \right],
        \label{eq:tvd_grad_final}
    \end{align}
    where \eqref{eq:A:2} follows from using the log-derivative trick. 
    Note that \label{eq:tvd_grad_final} is equivalent to the policy gradient where a policy and a reward are $p_\theta(x)$ and $r(x):=\mathbb{I}\{q(x) > p_{\theta}(x)\}$, respectively~\citep{williams1992simple}.
\end{proof}

\subsection{Model configurations}
\label{subsec:model_config}
The following configurations are used for our target and draft models:
\begin{table}[h]
    \centering
    \caption{Draft and target model configurations}
    \begin{tabular}{l||r|r}
    
           & Llama 2-Chat (7B, target) & Llama 2-Chat-Drafter (115M, draft) \\
           \hline
    Layers & 32 & 4\\
    Attention heads & 32 & 8\\
    Intermediate dim & 11,008 & 2,816\\
    Hidden dim & 2,048 & 1,024\\
    Activation & SiLU & SiLU
    \end{tabular}
    \label{tab:model_config}
\end{table}

\subsection{Training hyper-parameters}
For draft model pretraining, we used deepspeed stage1 with a batch-size=$496$ on 32 A100 GPUs. Additionally, during pre-training a large batch-size can be used as compared to distillation which requires the target model weight and output consuming a lot of memory. The optimizer used is \textit{AdamW} with  \textit{WarmUpDecayLR} learning-rate scheduler, maximum learning rate was $0.0001$ while minimum was $1e-6$, with $5000$ warm-up steps

For draft model fine-tuning, we used deepspeed stage $3$ with batch-size=$40$ on $8$ A-100 GPUs with maximum learning-rate=$0.0003$ with same optimizer and scheduler with $2000$ warm-up steps. 

\subsection{Data processing}
We preprocess the text data with the tokenizer of the target language model while appending End-Of-Sentence (EOS) token at the end of each input sequence. Furthermore, as a postporcessing step, all the sequences are concatenated into chunks of 2048 length, to maximize training throughput without adding pad tokens.

\subsection{Performance Degradation at Out-Of-Distribution (OOD) Task}
\begin{wrapfigure}{r}{0.4\textwidth}
    \vspace{-0.3in}
    \begin{center}
        \includegraphics[width=0.4\textwidth]{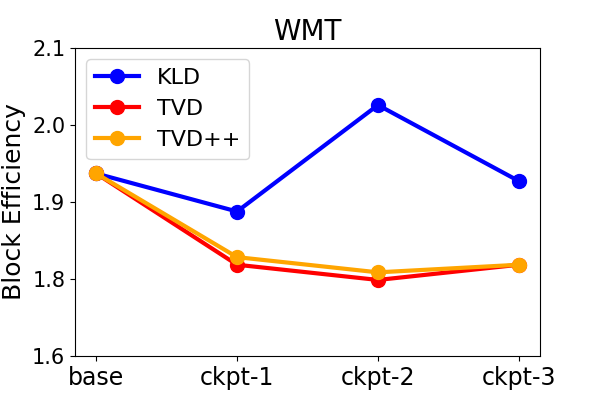}
    \end{center}
    \caption{Block efficiency for WMT18-DeEn results with multiple draft models are described.}
\end{wrapfigure}
We also evaluated the block efficiency of multiple draft models on translation task, WMT18-DeEn~\citep{bojar-EtAl:2018:WMT1}, from German to English. We found that all fine-tuned draft models were outperformed by the base model for this particular task since those models were \emph{not} directly fine-tuned on the translation task, making the task OOD.
We expect this issue will be resolved by adding in-distribution samples for training.


\end{document}